\documentclass{article}

\usepackage{arxiv}

\usepackage[utf8]{inputenc} 
\usepackage[T1]{fontenc}    
\usepackage{hyperref}       
\usepackage{url}            
\usepackage{booktabs}       
\usepackage{amsfonts}       
\usepackage{nicefrac}       
\usepackage{microtype}      
\usepackage{lipsum}
\usepackage{graphicx}
\usepackage[table]{xcolor}
\graphicspath{ {./images/} }

\title{Startup success prediction and VC portfolio simulation using CrunchBase data}

\author{
 Mark Potanin\thanks{Corresponding author}\\
  \texttt{potanin.m.st@gmail.com} \\
   \And
  Andrey Chertok \\
  \texttt{a.v.chertok@gmail.com} \\
   \And
  Konstantin Zorin \\
  \texttt{berzqwer@gmail.com} \\
     \And
  Cyril Shtabtsovsky \\
  \texttt{cyril@aloniq.com} \\
}

\begin{document}
\maketitle
\begin{abstract}

Predicting startup success presents a formidable challenge due to the inherently volatile landscape of the entrepreneurial ecosystem. The advent of extensive databases like Crunchbase jointly with available open data enables the application of machine learning and artificial intelligence for more accurate predictive analytics. This paper focuses on startups at their Series B and Series C investment stages, aiming to predict key success milestones such as achieving an Initial Public Offering (IPO), attaining unicorn status, or executing a successful Merger and Acquisition (M\&A). We introduce novel deep learning model for predicting startup success, integrating a variety of factors such as funding metrics, founder features, industry category. A distinctive feature of our research is the use of a comprehensive backtesting algorithm designed to simulate the venture capital investment process. This simulation allows for a robust evaluation of our model's performance against historical data, providing actionable insights into its practical utility in real-world investment contexts. Evaluating our model on Crunchbase's, we achieved a 14 times capital growth and successfully identified on B round high-potential startups including Revolut, DigitalOcean, Klarna, Github and others. Our empirical findings illuminate the importance of incorporating diverse feature sets in enhancing the model's predictive accuracy. In summary, our work demonstrates the considerable promise of deep learning models and alternative unstructured data in predicting startup success and sets the stage for future advancements in this research area.

\end{abstract}


\section{Introduction}

The prediction of startup success is a crucial task for various stakeholders, including investors, entrepreneurs, and policymakers, as it has significant implications for resource allocation and decision-making. It is estimated that approximately 90\% of startups fail within their first five years, a failure rate that has remained relatively constant over the past few decades, despite considerable advancements in technology and business practices. Consequently, the accurate prediction of startup success can assist investors in more effectively allocating their resources and enable entrepreneurs to make better-informed decisions.

Recently, the proliferation of data from sources such as Crunchbase has intensified interest in the application of machine learning techniques for the prediction of startup success. Machine learning models can harness various types of data, encompassing funding history, market trends, team composition, and social media activity, to identify patterns and generate predictions.


This study presents two distinct methodologies for predicting startup success: a supervised deep learning approach leveraging multiple data sources, and a ranking-based approach focusing on the identification of characteristics common to successful startups and investors. The supervised approach entails collecting and labeling data, constructing a prediction model, and evaluating its performance. In contrast, the ranking-based approach centers on identifying startups and investors that exhibit shared characteristics with successful ones.

Our train dataset consists of 34,470 companies The primary novelty of this research lies in the application of deep learning techniques and the integration of heterogeneous input data types. A crucial feature of our research is the simulation of fund operations based on historical data, resulting in a projected 14x capital growth of the fund's portfolio. As per machine learning metrics, our model exhibits a robust 86\% ROC\_AUC.

The remainder of this paper is organized as follows: Section 2 reviews the related works in the area of startup success prediction and machine learning. Section 3 describes dataset collection, preprocessing, and feature selection. Section 4 presents the experimental results of the supervised approach. Section 5 describes some other ideas about company and investor scoring. Finally, Sections 6 and 7 provide the conclusion of the study and discuss prospective research avenues in this domain.
\section{Related works}
The application of AI in fintech has substantially transformed the financial services industry over the past decades \cite{ahmed2022artificial}. For example, one of the most well-known applications is credit risk assessment \cite{shi2022machine}. Another challenging task could be stock market prediction \cite{jorgenson2023can}. This paper focuses on startup prediction and the VC market, and there is a growing literature on analyzing investments using machine learning.

In the paper \cite{ross2021capitalvx}, authors present a machine learning model, CapitalVX, trained on a large dataset obtained from Crunchbase, to predict the outcomes for startups, i.e., whether they will exit successfully through an IPO or acquisition, fail, or remain private. They investigated MLP, Random Forest, XGBoost and used mostly numerical features from the dataset. In \cite{10.1145/3484824.3484919} paper, authors conducted a review on existing machine learning techniques that are recently contributed to understanding the need of start-ups, trends of business and can provide recommendations to plan their future strategies to deal with the business problems. The study conducted by \cite{arroyo2019assessment} underscores the potential of machine learning applications in the venture capital industry, demonstrating its ability to predict various outcomes for early-stage companies including subsequent funding rounds or closure. 

In another study \cite{blohm2022sa}
, authors use behavioral decision theory to compare the investment returns of an algorithm with those of 255 business angels (BAs) investing via an angel investment platform. The study found that, on average, the algorithm achieved higher investment performance than the BAs. However, experienced BAs who were able to suppress their cognitive biases could still achieve best-in-class investment returns. This research presents novel insights into the interplay of cognitive limitations, experience, and the use of algorithms in early-stage investing. This study \cite{corea2021hacking} proposes a data-driven framework, wherein the model was trained on 600,000 companies across two decades and 21 significant features. 

This review \cite{corea2019ai} provides a thorough analysis of AI applications in Venture Capital, categorizing influential factors on a company's probability of success or fund-raising into three clusters: team/personal characteristics, financial considerations, and business features. In another study \cite{zbikowski2021machine}, authors leveraged Crunchbase data from 213,171 companies to develop a machine learning model to predict a company's success. Despite limiting the number of predictors, it achieved promising results in precision, recall, and F1 scores, with the best outcomes from the gradient boosting classifier. 

This study \cite{sharchilev2018web} explores the untapped potential of web-based open sources in contrast to just structured data from the startup ecosystem. A significant performance enhancement is demonstrated by incorporating web mentions of the companies into a robust machine learning pipeline using gradient boosting.



This study \cite{singhal2022data} aims to assist VC firms and Angel investors in identifying promising startups through rigorous evaluations, emphasizing the impact of founder backgrounds and capital collected in seed and series stages.  
This very recent paper published in 2023 \cite{kim2023succeed} introduces a novel model for predicting startup success that incorporates both internal conditions and industry characteristics, addressing a gap in previous research that focused primarily on internal factors. Using data from over 218,000 companies from Crunchbase and six machine learning models, the authors found media exposure, monetary funding, the level of industry convergence, and the level of industry association to be key determinants of startup success.

In this study \cite{antretter2019predicting}, authors analyze more than 187,000 tweets from 253 new ventures’ Twitter accounts achieving up to 76\% accuracy in discriminating between failed and successful businesses . 
The research outlined in \cite{tumasjan2021twitter} investigates the methodologies used by venture capitalists when evaluating technology-based startups, using the influence of weak (Twitter sentiment) and strong (patents) signals on venture valuations. Findings reveal that while both signals positively associate with venture valuations, Twitter sentiment fails to correlate with long-term investment success, unlike patents. Furthermore, startup age and VC firm experience act as boundary conditions for these signal-valuation relationships.


\section{Dataset Overview, Preprocessing, and Features}
\label{sec:headings}

We used daily Crunchbase database export (Daily CSV Export) as the primary data source, which is also supported by a well-documented API. The main goal of this research was to collect a labeled dataset for training a deep learning model to classify companies as either successful or unsuccessful. 

The analysis was based on the Daily CSV Export from 2022-06-14, and only companies established on or after 2000-01-01 were taken into account. To refine the focus of the research, only companies within specific categories were included, such as \textit{Software}, \textit{Internet Services}, \textit{Hardware}, \textit{Information Technology}, \textit{Media and Entertainment}, \textit{Commerce and Shopping}, \textit{Mobile}, \textit{Data and Analytics}, \textit{Financial Services}, \textit{Sales and Marketing}, \textit{Apps}, \textit{Advertising}, \textit{Artificial Intelligence}, \textit{Professional Services}, \textit{Privacy and Security}, \textit{Video}, \textit{Content and Publishing}, \textit{Design}, \textit{Payments}, \textit{Gaming}, \textit{Messaging and Telecommunications}, \textit{Music and Audio}, \textit{Platforms}, \textit{Education}, and \textit{Lending and Investments}.


This research is focused on investment rounds occurring after round B. However, in the Crunchbase data glossary, rounds such as \textit{series\_unknown}, \textit{private\_equity}, and \textit{undisclosed}, possess unclear characteristics. To incorporate them into the company’s funding round history, we only included these ambiguous rounds if they occurred after round B.

\subsection{Successful Companies Dataset}
In this research, a company is deemed successful if it achieves one of three outcomes: Initial Public Offering (IPO), Acquisition (ACQ), or Unicorn status (UNIC), the latter being defined as a valuation exceeding \$1 billion. To assemble a list of successful companies, we initially filtered for IPOs with valuations above \$500M or funds raised over \$100M, yielding 363 companies. For acquisitions, we applied filters to eliminate companies with a purchase price below the maximum amount of funds raised or under \$100M, resulting in 833 companies. To select unicorns, we searched for companies with a valuation above \$1 billion, utilizing both Crunchbase data and an additional table of verified unicorns, which led to a total of 1074 unicorns.

The final dataset contains a timeline of all crucial investment rounds leading to the success event (i.e., achieving unicorn status, IPO, or ACQ), with the index of this event specified in the \textit{success\_round} column. This approach ensures that the dataset accurately represents the history and progress of each successful company, facilitating effective analysis.

\subsection{Unsuccessful Companies Dataset}
To supply the model with examples of 'unsuccessful' companies, we collected a separate dataset. We excluded companies already present in the successful companies dataset by removing those that had IPO, ACQ, or UNIC flags. We also eliminated a considerable number of actual unicorns from the CrunchBase website \cite{crunchbase} to avoid overlap. We excluded companies that have not attracted any rounds since 2016. Additionally, we excluded companies that are subsidiaries or parent companies of other entities. Furthermore, we used the jobs dataset to exclude companies that have hired employees since 2017. 

Additionally, we applied extra filters to exclude companies with valuation above \$100 million, as they reside in the "gray area" of companies that may not be clearly categorized as successful or unsuccessful. 
By applying these filters, we constructed a dataset comprising 32,760 companies, denoted by the label '0' for unsuccessful, and 1,989 companies, denoted by the label '1' for successful. 

\subsection{Features}

The feature space of the model includes:

\paragraph{Founders Features}\
Categorical: \textit{country\_code}, \textit{region}, \textit{city}, \textit{institution\_name}, \textit{degree\_type}, \textit{subject}.\\
Numerical: \textit{twitter\_url}, \textit{linkedin\_url}, \textit{facebook\_url}, \textit{gender}, \textit{is\_completed}, \textit{num\_degrees}, \textit{num\_last\_startups}, \textit{num\_last\_jobs}, \textit{number\_of\_founders}.

We incorporated three binary flags into our model to represent the presence of founders’ social media links. Since a company can have multiple founders, it was essential to aggregate information on all the founders for each company. For categorical variables, the most frequent value from the list was used, and the median for numerical variables.

\paragraph{Investors Features}

Categorical: \textit{type}, \textit{country\_code}, \textit{region}, \textit{city}, \textit{investor\_types}.\\
Numerical: \textit{investment\_count}, \textit{total\_funding\_usd}, \textit{twitter\_url}, \textit{linkedin\_url}, \textit{facebook\_url}, \textit{raised\_amount\_usd}, \textit{investor\_count}, \textit{num\_full}.

Functions that generate features based on the founders' and investors' data incorporate a date parameter as input. This approach is necessary to the model from using future information. For example, details about the number of companies founded or the founder's previous job experience that took place after the date of interest should not be incorporated into the feature set to avoid information leakage from the future.

\paragraph{Rounds features}

Categorical: \textit{country\_code}, \textit{investment\_type}, \textit{region}, \textit{city}, \textit{investor\_name}.\\
Numerical: \textit{sum}, \textit{mean}, \textit{max} of \textit{raised\_amount\_usd}, \textit{investor\_count}, \textit{post\_money\_valuation\_usd}.

It is crucial to emphasize that all features related to a company’s investment rounds are gathered at a time point prior to the beginning of the time window of interest. 



\paragraph{Categories}

There are two additional types of text data - text tags representing the companies' field of work. For example:

\begin{itemize}
    \item category\_list: \textit{Internet, Social Media, Social Network}
    \item category\_groups\_list: \textit{Data and Analytics, Information Technology, Software}
\end{itemize}

The set of tags used in our study consists of a list of keywords separated by commas. We used the NMF (Non-Negative Matrix Factorization) matrix factorization method to generate features from these tags. This process involves creating a binary table with companies represented as rows and tags as columns, where each value in the table indicates whether a given company is associated with a specific tag (1) or not (0). The trained matrix factorization then converts each binary vector into a smaller dimension vector (in our case, 30).

All categorical features are encoded using the OrdinalEncoder, while numerical features are normalized.

\section{Model Training, Evaluation, and Portfolio Simulation}

\begin{figure} 
    \centering
    \caption{Model architecture}
    \label{fig:model}
    \includegraphics[scale=0.08]{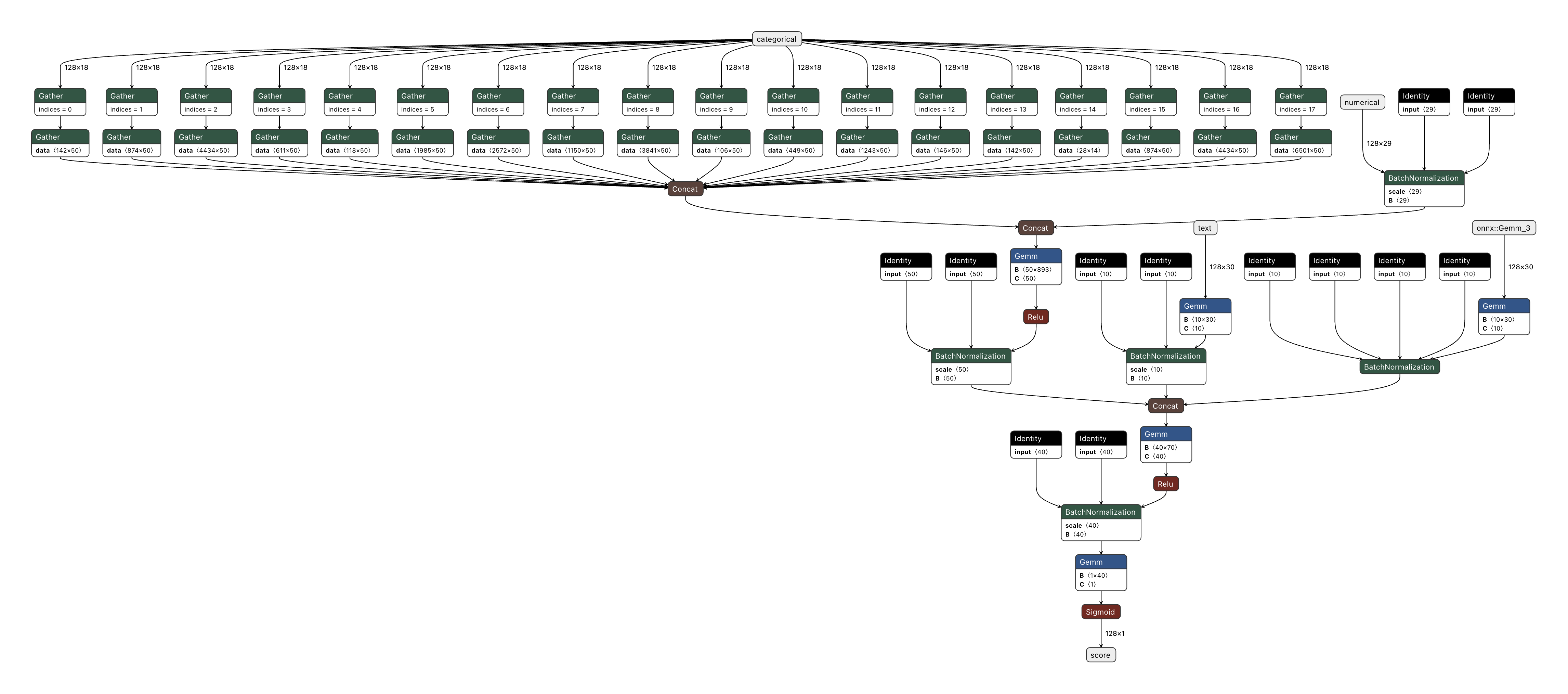}
\end{figure}

A representation of the model's architecture is visualized in Figure \ref{fig:model}.

\subsection{Backtest}

The backtest period during which we tested the model spans from 2016-01-01 to 2022-01-01, and the model was retrained every 3 months. Actually, it is a hyperparameter that could be tuned depending on the time/accuracy trade-off. In each iteration of the backtest, the time window under consideration is defined by the start and end dates. For example, the first iteration considers the window with a start date of 2016-01-01 and an end date of 2016-04-01. Companies that attracted Round B or C are selected as the "test" set for this window.

The model is trained on the dataset described in Section 3. However, the entire dataset cannot be used for training since it would be incorrect to train on companies founded in the future to predict the success of companies in the past. Therefore, only companies founded before the start of the current time window (i.e., before 2016-01-01 in the first iteration) are considered for training. Additionally, the success of a company (IPO/ACQ/UNICORN) may occur in the future relative to the current window. To train the model, only companies with success event occurred before the start of the current time window are considered.

This approach is designed to ensure the integrity of the backtesting process, avoiding any influence from future events. However, the drawback of this approach is the limited number of training examples at the beginning of the backtest (i.e., in the first iterations in 2016-2017). Consequently, the predictive power of the model is lower at the beginning of the backtest compared to the end. The backtest yields an array of test companies with a score assigned to them, indicating the level of success predicted by the model. 

The model is retrained every 3 months during the backtest, resulting in a total of 25 prediction windows. A sorted list of predictions is generated for each window. Finally, all predictions from all windows are compiled into one table, representing the complete backtest of predictions for the period from 2016-01-01 to 2022-01-01. This table passes to the optimization algorithm.

A decision has been made to construct a monthly portfolio based on the backtest results. Therefore, we can conduct the backtest with a window of 1 month, covering the periods from 2016-01-01 to 2016-02-01, from 2016-02-01 to 2016-03-01, and so on, by adding or removing companies in our portfolio every month. Again, using the example of one period, for example from 2018-01-01 to 2018-02-01, let us describe the process of selecting companies to be included in the portfolio. We take a slice of the backtest predictions in this period and sort them by score, which represents the model's assessment of the success of each company. As the size of our portfolio is limited, for instance, to 30 companies, there is no need to fill it entirely in the first months. Thus, the logic for adding companies is as follows:

\begin{itemize}
   \item In each month, we select the top 3 companies from the sorted list of predictions. But, a cut-off threshold for the predicted score has also been established. However, the choice of the optimal threshold is an empirical task and requires careful consideration. With the augmentation of the training dataset over time, the model becomes more confident in its predictions. Therefore, it makes sense to increase the threshold when moving along the backtest time. One way to do this is to set the threshold as a function that takes into account the size of the train dataset and other relevant factors.
   \item Every month we verify the current portfolio:
   \begin{itemize}
     \item \textbf{success}: if the company has achieved a success event (IPO/ACQ/unicorn) during the month, it is removed from the active portfolio and marked with this flag.
     \item \textbf{longtime}: if the company has not attracted any rounds within the last 730 days (2 years, a configurable parameter), it is removed from the portfolio and marked with this flag.
     \item \textbf{still\_in}: if the company is still in the portfolio at the end of the backtest, it is marked with this flag. These are companies that were recently added to the portfolio (in 2021-2022) and for which we cannot yet make a decision on their success.
    \end{itemize}
 \end{itemize}

The result is a dataset that simulates our venture fund during the period 2016-2022, and we collected (as well as filtered) companies in it every month. The resulting dataset contains the following fields:

\begin{itemize}
\item uuid - a unique company identifier
\item name - the name of the company
\item enter\_series\_date - the date of the round in which the fund entered the company
\item enter\_series\_value - the valuation of the company at the time of entry (if available)
\item score - the company score at the time of entry (if available)
\item added - the date when the company was added to the portfolio
\item last\_series\_date - the date of the last round of funding, which could be an IPO, acquisition, or the round in which the company became a unicorn
\item last\_series\_value - the valuation of the company at the time of the last round of funding, if available
\item exit\_reason - the reason for the fund's exit from the company (if applicable)
\item expired - the date when the fund exited the company (due to success or expiration of the holding period)
\end{itemize}

The reader may wonder why we retrain the model every 3 months while building the portfolio with a one-month interval. Essentially, at the beginning of the training set, we include all companies until 2016-01-01. The test set consists of companies that received rounds B or C funding during the period from 2016-01-01 to 2016-04-01. We make predictions and add them to the overall table. Then, we expand the training data until 2016-04-01, and the test period becomes from 2016-04-01 to 2016-07-01, and so on. In the end, we have a complete test table covering the period from 2016-01-01 to 2022-01-01.

After that, we go through this table with a one-month step, simulating our venture fund's behavior and assembling the portfolio. The fact that we first collect all predictions and then go through them to construct the portfolio is simply a matter of optimization. We do not look into the future in any way.

\subsection{Backtest settings}

In this study, several experiments were conducted with different backtest configurations, we called them \textbf{earlybird} and \textbf{any}. The earlybird configuration exclusively permits entry for companies only in rounds B or C, while the any configuration broadens the entry criteria to any round within the list of \textit{series\_b},\textit{series\_c},\textit{series\_d},\textit{series\_e},\textit{series\_f}, \textit{series\_g},\textit{series\_h},\textit{series\_j},\textit{series\_i}, as long as they are within the considered backtest window.

The choice of entry configuration depends on the stage at which we enter the company. Similarly, the choice of exit configuration depends on when we decide to exit the company based on its success event (IPO/ACQ/unicorn), as discussed in Section 2.2. However, since the "unicorn" status can occur in the early rounds, there is a question of which round to exit. Two approaches were considered: using \textbf{first} approach we exit the company when the first success event occurs, while using \textbf{last} approach we exit on the last success event, analogous to "we sit until the end."

The main approach used in this study is the \textbf{earlybird\_last} due to business requirements. However, this approach has its drawbacks, such as the fact that the company success flag becomes known later in time, resulting in a smaller dataset size for training at the beginning of the backtest and a slightly lower quality of the backtest compared to the \textbf{earlybird\_first} approach.

\subsection{Results}

The primary output of the algorithm is the backtest Table \ref{table:earlybird_last}, sorted by the time the company was added to the portfolio. The table includes an \textit{exit\_reason} column, which serves as the main metric for evaluating model quality on the backtest. This column can take on the following values:
\begin{itemize}
\item \textbf{success}: the company had a successful round (unicorn/acquisition/IPO), and we exited
\item \textbf{longtime}: a negative case where we left the company because it didn't have a successful event and had no rounds for two years
\item \textbf{STILL\_IN}: a gray area, mainly consisting of companies that were recently added to the backtest
\end{itemize}

Hence, an optimal backtest is characterized by the maximum quantity of successful companies and a minimal number of companies categorized as \textbf{longtime}. Table \ref{table:earlybird_last} (\textit{earlybird\_last}) is the basic configuration based on business requirements. We enter in the first rounds (B/C) and exit in the last round. However, the model may not work very well at the beginning of the backtest due to limited data for training. In the Table \ref{table:any_last} (\textit{any\_last}) configuration, we can observe a large number of known unicorns simply because we allow the model to enter in later rounds.

\subsection{Capital Growth}

Traditional metrics utilized in machine learning may not be directly transferable to the AI investor due to changes in data availability over time and class imbalance in the dataset. Therefore, we assess the model's performance based on the presence of well-known companies in the resulting portfolio and the financial growth of the companies. In this subsection, we focus on the latter assessment.

To calculate the PnL of the success of companies, we need the company valuation during entry and exit rounds. The valuation of companies that exited due to longtime is set to zero. For companies marked as \textbf{STILL\_IN}, we use their last known valuation since they are the youngest companies in the portfolio. The PnL is divided into realized and unrealized components. The unrealized PnL illustrates the current cumulative valuation of the portfolio, incorporating the presently known rounds, in contrast, the realized PnL denotes the cumulative sum garnered by exiting thriving companies and consequent capital growth.
Results with exit reasons and valuations are presented in Table \ref{table:earlybird_last}.  Unfortunately, we didn't have valuation data for all companies. There is a column 'Used in Capital Growth' that shows whether the company was used to calculate the PnL.

\begin{figure} 
    \centering
    \caption{Capital growth}
    \label{fig:pnl}
    \includegraphics[scale=0.35]{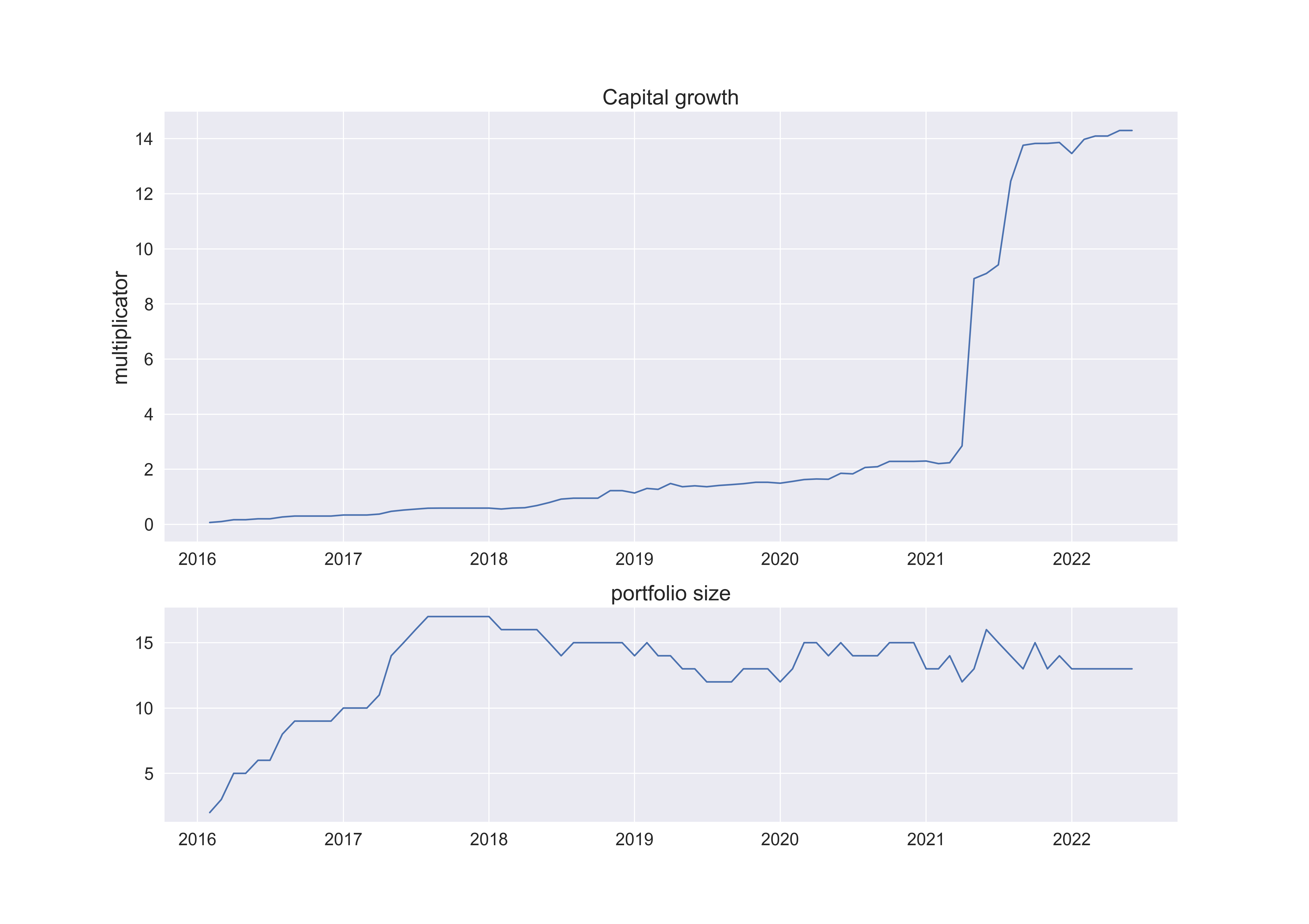}
\end{figure}

We present cumulative PnL and the current portfolio size over time in Figure \ref{fig:pnl}, with a step size of 1 month. The sharp rise in the middle of 2021 corresponds to the exit from Revolut. The companies that remained in the portfolio at the end of 2021 are all marked as \textbf{STILL\_IN}. Overall, the PnL graph shows a positive trend, indicating the financial growth of the portfolio over time. 

To evaluate the algorithm via conventional machine learning metrics, we employ cross-validation for time-series analysis with a 1-year test window, spanning the years from 2016 to 2022. Within this test window, we focus on companies that secured B or C funding rounds during a given year and subsequently achieved success. 
Furthermore, to ensure the integrity of our analysis, the training dataset for each fold exclusively comprises companies whose success or failure status was known prior to the commencement of the test window.  Standard binary classification metrics can be used to evaluate the performance of the model, and Recall is of particular interest to us. The minimization of False Negatives (FN) holds greater significance than that of False Positives (FP) in order to circumvent the omission of successful companies. Finally, in Table \ref{table:metrics} we present metrics that have been averaged across 6 folds for a comprehensive evaluation of our predictive model's performance:
\begin{table}[htbp]
    \centering
    \caption{Metrics}
    \label{table:metrics}
\begin{tabular}{|c|c|c|c|}
\hline
\textbf{Precision} & \textbf{Recall} & \textbf{ROC AUC} & \textbf{PR AUC} \\
\hline
0.92 & 0.64 & 0.86 & 0.65 \\
\hline
\end{tabular}
\end{table}

\section{Other approaches}

\subsection{Investors ranking model}

All investors could be scored in terms of frequency, amount, and field of investments. Also, an investor could be an indicator of a company’s potential failure or success. This scoring was carried out in three stages:
\begin{enumerate}
\item  Through an autoencoder model with several modalities, we created vector representations for each investor
\item According to experts’ estimates, we select a group of top investors, and further create the centroid of this group in the vector space
\item We rank investors according to distance from the centroid
\end{enumerate}

An elevated score corresponds to a proximate alignment with top investors. Results are presented in Table \ref{tab:investor_scores}. If the lead investor of a company has a low score, it could be an indicator that such a company should be excluded from consideration.

\textbf{Example:} Company 14W has a score of 0.9 and invests in IT companies, incl. unicorns (for example, European travel management startup TravelPerk).

\subsection{Founders ranking model}

According to some characteristics - the number of previous startups (founder, co-founder), their area, success, etc. - we can also score founders. An escalated score is indicative of a company's enhanced credibility. The results of these models can be used both for preliminary scoring of companies and as independent features in other models. An example is presented in Table \ref{tab:founder_scores}.


\subsection{Unicorn recommendation model}

It was revealed that the median time for a company to achieve the status of a "unicorn" is 4-5 years. Thus, in this period of time, about half of the unicorns have reached this status, moreover, the second half is waiting in the wings in the near future. This model identifies nascent companies established within this 4-5 year time frame, isolates 'unicorns' within this subset, scores entities bearing the greatest resemblance and subsequently generates a list of the top 30 recommendations.

For 2016-2021 simulation run:
\begin{itemize}
\item On Jan 1st of each year, a list of recommendations of potential unicorns is formed.
\item Every month, in case of the announcement of a round (series\_X), a company is added to the portfolio if its valuation is below 1 billion and the round is not too high.
\item Companies that have reached 2.5 billion or have not had rounds for 3 years are removed from the portfolio.
\end{itemize}

As a result, at the end of the period, a portfolio of companies was formed. The limitation in this context is the scarcity of information related to {post\_money\_valuation} information. Further development: as new data become available, building a more complex recommendation system. The results are presented in Table \ref{tab:unicorn_scoring}.

\section{Conclusion}
Traditionally, venture capital investment decisions have largely been guided by the investors' intuition, experience, and market understanding. While these elements remain significant, there's a growing recognition that these traditional approaches can be greatly enhanced by integrating data-driven insights into the investment decision-making process. 

Our paper comprehensively examines a predictive model for startups based on an extensive dataset from CrunchBase. A meticulous review and analysis of the available data were conducted, followed by the preparing of a dataset for model training. Special attention was given to the selection of features which include information about founders, investors, and funding rounds. 

The article also underlines a thoughtfully designed backtest algorithm, enabling a fair evaluation of the model's behavior (and the simulation of a VC fund based on it) from a historical perspective. Rigorous efforts were made to avoid data leakage, ensuring training at any given point only utilized data that would have been known at that time. Several configurations were explored regarding the funding rounds at which the fund could invest in a company and the timing of exits. The primary evaluative metrics were derived from a backtest table (Table \ref{table:earlybird_last}), which chronicles instances of company entries, exits, and the corresponding success statuses. Utilizing additional data on company valuations, we calculated the Capital Growth, illustrating the fund's impressive economic impact over time. To sum up, this work primarily focused on the variety of input features, the integrity of the backtest, and the realistic simulation of the portfolio from a historical perspective. Additionally, we proffer a series of propositions aimed at enhancing the existing model, primarily revolving around the access to supplementary data repositories.



Within the highly competitive and dynamic investment environment, the assimilation of data-driven decision-making practices transitions from being an option to becoming a necessity. As such, venture capitalists that effectively harness the potential of AI and machine learning will likely secure a significant competitive advantage, positioning themselves for success in the new era of venture capitalism.

\section{Further Research}

In terms of further work, a promising direction is the usage of different sources of text data about companies, founders, and investors. This could involve leveraging social media platforms such as Twitter and LinkedIn, as well as parsing the websites of the companies themselves.

Additionally, it may be worth adjusting the foundation date filter to include companies founded in 1995, rather than the current start date of 2000-01-01. However, this could potentially result in an influx of companies from the "dotcom bubble" period.


The current strict filters used to determine successful companies (IPO/ACQ/UNICORN) could also be loosened to potentially capture more companies in the "gray area" between success and failure.

Finally, it may be worth conducting experiments to determine the optimal threshold value for adding companies to the portfolio, taking into account the size of the portfolio.

These additional tasks can provide valuable insights and enhance the effectiveness of the AI investor backtest model. Analyzing the presentation materials, video interviews, and source code of software companies can provide a better understanding of the company's strategy, goals, and potential. Developing information collection systems to automate this process can save time and improve accuracy.

Evaluating the influence of macroeconomic elements and technological trajectories on startups may facilitate the identification of potential risks and opportunities. It can also aid in the development of exit strategies. Additionally, analyzing competing studies can provide insights into the market and competition, which can inform investment decisions.


\bibliographystyle{unsrt}  
\bibliography{references}  

\appendix
\section{Appendix A}
\begin{table}[htbp]
    \centering
    \caption{Earlybird Last}
    \label{table:earlybird_last}
    \scriptsize
    \begin{tabular}{|c|c|c|c|c|c|}
        \hline
        \textbf{Name} & \textbf{Added} & \textbf{Exit Reason} & \textbf{Expelled}& \textbf{Used in Capital Growth} \\
        \hline
        Flatiron & 2016-02-01 & longtime & 2018-02-01 & TRUE \\
        \rowcolor{green!25}
Klarna & 2016-02-01 & success & 2021-07-01 & TRUE \\
\rowcolor{green!25}
Looker & 2016-02-01 & success & 2019-07-01 & FALSE \\
\rowcolor{green!25}
Magic Leap & 2016-03-01 & success & 2021-11-01 & TRUE \\
Jumia Group & 2016-04-01 & longtime & 2018-04-01 & TRUE \\
Monedo & 2016-04-01 & longtime & 2019-01-01 & TRUE \\
\rowcolor{green!25}
Vectra & 2016-04-01 & success & 2021-05-01 & FALSE \\
Pivotal & 2016-06-01 & longtime & 2018-06-01 & TRUE \\
Barefoot Networks & 2016-07-01 & longtime & 2020-09-01 & FALSE \\
CommonBond & 2016-08-01 & longtime & 2020-04-01 & FALSE \\
\rowcolor{green!25}
GitHub & 2016-08-01 & success & 2018-07-01 & TRUE \\
\rowcolor{green!25}
Coinbase & 2016-08-01 & success & 2021-05-01 & TRUE \\
\rowcolor{green!25}
Auris Health & 2016-09-01 & success & 2019-03-01 & TRUE \\
Distil Networks & 2016-09-01 & longtime & 2018-09-01 & FALSE \\
Behalf & 2016-09-01 & STILL IN & & FALSE \\
\rowcolor{green!25}
PillPack & 2017-01-01 & success & 2018-07-01 & FALSE \\
Pluralsight & 2017-01-01 & longtime & 2019-01-01 & TRUE \\
\rowcolor{green!25}
Exabeam & 2017-03-01 & success & 2021-06-01 & FALSE \\
Learner-Centered Collaborative & 2017-04-01 & longtime & 2020-05-01 & FALSE \\
\rowcolor{green!25}
Earnest Operations & 2017-04-01 & success & 2017-11-01 & FALSE \\
Aledade & 2017-04-01 & longtime & 2020-01-01 & TRUE \\
Qualtrics & 2017-05-01 & longtime & 2019-05-01 & TRUE \\
\rowcolor{green!25}
Cohesity & 2017-05-01 & success & 2020-05-01 & TRUE \\
\rowcolor{green!25}
Robinhood Markets & 2017-05-01 & success & 2021-08-01 & TRUE \\
Wandera & 2017-06-01 & longtime & 2019-06-01 & FALSE \\
OpenGov & 2017-06-01 & longtime & 2021-10-01 & FALSE \\
\rowcolor{green!25}
Symphony Communication Services & 2017-06-01 & success & 2019-07-01 & TRUE \\
Casper & 2017-07-01 & longtime & 2021-04-01 & TRUE \\
\rowcolor{green!25}
Segment & 2017-08-01 & success & 2020-11-01 & FALSE \\
\rowcolor{green!25}
Revolut & 2017-08-01 & success & 2021-08-01 & TRUE \\
\rowcolor{green!25}
Spire Global & 2017-12-01 & success & 2021-09-01 & FALSE \\
ClearStory Data & 2018-04-01 & longtime & 2020-04-01 & TRUE \\
Helix & 2018-04-01 & longtime & 2020-03-01 & FALSE \\
\rowcolor{green!25}
Aircall & 2018-06-01 & success & 2021-07-01 & FALSE \\
\rowcolor{green!25}
Zoox & 2018-08-01 & success & 2020-07-01 & TRUE \\
Yesware & 2018-09-01 & longtime & 2020-09-01 & FALSE \\
import.io & 2019-01-01 & longtime & 2021-01-01 & FALSE \\
Fair & 2019-01-01 & longtime & 2021-01-01 & TRUE \\
Petal & 2019-02-01 & STILL IN & & TRUE \\
Bolt & 2019-08-01 & STILL IN & & FALSE \\
\rowcolor{green!25}
Ginger & 2019-10-01 & success & 2021-09-01 & TRUE \\
Blue Planet-works Japan & 2019-10-01 & longtime & 2021-10-01 & FALSE \\
\rowcolor{green!25}
Porch Group & 2020-02-01 & success & 2021-01-01 & TRUE \\
\rowcolor{green!25}
OneTrust & 2020-03-01 & success & 2021-05-01 & TRUE \\
Pony.ai & 2020-03-01 & STILL IN & & TRUE \\
\rowcolor{green!25}
Bakkt & 2020-04-01 & success & 2021-11-01 & TRUE \\
\rowcolor{green!25}
DigitalOcean & 2020-06-01 & success & 2021-04-01 & TRUE \\
Wandelbots & 2020-07-01 & STILL IN & & FALSE \\
Withings & 2020-08-01 & STILL IN & & FALSE \\
\rowcolor{green!25}
Northvolt & 2020-10-01 & success & 2021-07-01 & TRUE \\
WireWheel & 2021-03-01 & STILL IN & & TRUE \\
Skycatch & 2021-04-01 & STILL IN & & FALSE \\
Bitstocks Trading Limited & 2021-05-01 & STILL IN & & TRUE \\
Placer.ai & 2021-05-01 & STILL IN & & TRUE \\
Grover & 2021-05-01 & STILL IN & & TRUE \\
Genies & 2021-06-01 & STILL IN & & TRUE \\
Heyday & 2021-06-01 & STILL IN & & TRUE \\
\rowcolor{green!25}
Pipe & 2021-06-01 & success & 2022-01-01 & TRUE \\
Lendbuzz & 2021-07-01 & STILL IN & & FALSE \\
ShipHawk & 2021-07-01 & STILL IN & & FALSE \\
DuckDuckGo & 2021-07-01 & STILL IN & & TRUE \\
Railsbank & 2021-08-01 & STILL IN & & TRUE \\
Ripio & 2021-10-01 & STILL IN & & TRUE \\
Jeeves & 2021-10-01 & STILL IN & & TRUE \\
slice & 2021-12-01 & STILL IN & & TRUE \\
Terra & 2022-01-01 & STILL IN & & FALSE \\
        \hline
    \end{tabular}
\end{table}

\begin{table}[htbp]
    \centering
    \scriptsize
    \caption{Any Last}
    \label{table:any_last}
    \begin{tabular}{|c|c|c|c|c|}
        \hline
        \textbf{Name} & \textbf{Enter Series} & \textbf{Added} & \textbf{Exit Reason} & \textbf{Expelled} \\
        \hline
Klarna & series\_c & 2016-02-01 & success & 2021-07-01 \\
Snapdeal & series\_j & 2016-02-01 & longtime & 2019-06-01 \\
\rowcolor{green!25} 
Magic Leap & series\_c & 2016-03-01 & success & 2021-11-01 \\
Carbon Black & series\_f & 2016-03-01 & longtime & 2018-03-01 \\
\rowcolor{green!25} 
Domo & series\_d & 2016-04-01 & success & 2018-07-01 \\
Monedo & series\_c & 2016-04-01 & longtime & 2019-01-01 \\
\rowcolor{green!25} 
Airbnb & series\_e & 2016-05-01 & success & 2021-01-01 \\
Slack & series\_f & 2016-05-01 & longtime & 2020-09-01 \\
\rowcolor{green!25} 
Snap & series\_f & 2016-06-01 & success & 2017-04-01 \\
Pivotal & series\_c & 2016-06-01 & longtime & 2018-06-01 \\
Anki & series\_d & 2016-07-01 & longtime & 2019-04-01 \\
\rowcolor{green!25}
Uber & series\_g & 2016-07-01 & success & 2019-06-01 \\
\rowcolor{green!25} 
GitHub & series\_b & 2016-08-01 & success & 2018-07-01 \\
Behalf & series\_c & 2016-09-01 & STILL IN & \\
Kaltura & series\_f & 2016-09-01 & longtime & 2018-09-01 \\
Sckipio Technologies & series\_c & 2016-10-01 & longtime & 2020-06-01 \\
\rowcolor{green!25} 
DraftKings & series\_e & 2016-10-01 & success & 2020-05-01 \\
Paxata & series\_d & 2016-11-01 & longtime & 2018-11-01 \\
BuzzFeed & series\_g & 2016-12-01 & longtime & 2018-12-01 \\
\rowcolor{green!25} 
Stripe & series\_d & 2016-12-01 & success & 2021-04-01 \\
Pluralsight & series\_c & 2017-01-01 & longtime & 2019-01-01 \\
Nexenta Systems & series\_g & 2017-01-01 & longtime & 2019-01-01 \\
\rowcolor{green!25} 
Funding Circle & series\_f & 2017-02-01 & success & 2018-10-01 \\
InsideSales & series\_d & 2017-02-01 & longtime & 2019-02-01 \\
\rowcolor{green!25} 
SoFi & series\_f & 2017-03-01 & success & 2021-06-01 \\
Comprehend Systems & series\_c & 2017-03-01 & longtime & 2021-03-01 \\
\rowcolor{green!25} 
Earnest & series\_c & 2017-04-01 & success & 2017-11-01 \\
\rowcolor{green!25} 
Instacart & series\_d & 2017-04-01 & success & 2021-04-01 \\
Flipkart & series\_j & 2017-05-01 & longtime & 2019-09-01 \\
\rowcolor{green!25} 
Lyft & series\_g & 2017-05-01 & success & 2019-04-01 \\
\rowcolor{green!25} 
Wish & series\_f & 2017-06-01 & success & 2021-01-01 \\
The Honest Company & series\_e & 2017-11-01 & longtime & 2020-07-01 \\
\rowcolor{green!25} 
Warby Parker & series\_e & 2018-04-01 & success & 2021-10-01 \\
Nextdoor & series\_g & 2018-06-01 & longtime & 2021-10-01 \\
CloudBees & series\_e & 2018-07-01 & longtime & 2020-07-01 \\
Dataminr & series\_e & 2018-07-01 & longtime & 2020-07-01 \\
Numerify & series\_d & 2018-11-01 & longtime & 2020-11-01 \\
import.io & series\_b & 2019-01-01 & longtime & 2021-01-01 \\
Fair & series\_b & 2019-01-01 & longtime & 2021-01-01 \\
Petal & series\_b & 2019-02-01 & STILL IN & \\
Ola & series\_j & 2019-02-01 & STILL IN & \\
Percolate & series\_d & 2019-04-01 & longtime & 2021-04-01 \\
\rowcolor{green!25} 
Cloudflare & series\_e & 2019-04-01 & success & 2019-10-01 \\
Optimizely & series\_d & 2019-07-01 & success & 2020-10-01 \\
Dstillery & series\_d & 2019-07-01 & longtime & 2021-07-01 \\
\rowcolor{green!25} 
Thumbtack & series\_f & 2019-08-01 & success & 2021-07-01 \\
Zocdoc & series\_d & 2019-08-01 & STILL IN & \\
\rowcolor{green!25} 
One97 & series\_g & 2019-12-01 & success & 2021-12-01 \\
ATLAS CAPITAL GROUP & series\_b & 2020-02-01 & STILL IN & \\
\rowcolor{green!25} 
OneTrust & series\_b & 2020-03-01 & success & 2021-05-01 \\
\rowcolor{green!25} 
DigitalOcean & series\_c & 2020-06-01 & success & 2021-04-01 \\
Wandelbots & series\_b & 2020-07-01 & STILL IN & \\
\rowcolor{green!25} 
Northvolt & series\_b & 2020-10-01 & success & 2021-07-01 \\
Splashtop & series\_e & 2021-02-01 & STILL IN & \\
Bitstocks Trading Limited & series\_c & 2021-05-01 & STILL IN & \\
\rowcolor{green!25} 
Pipe & series\_b & 2021-06-01 & success & 2022-01-01 \\
Ripio & series\_b & 2021-10-01 & STILL IN & \\
Delphix & series\_d & 2021-11-01 & STILL IN & \\
        \hline
    \end{tabular}
\end{table}

\newpage

\begin{table}
  \begin{minipage}{.5\linewidth}
    \centering
        \caption{Investors scoring}
    \label{tab:investor_scores}
    \scriptsize
    \begin{tabular}{|c|c|}
        \hline
        \textbf{Name} & \textbf{Score} \\
        \hline
        14W & 0.900773 \\
        Mubadala Capital Ventures & 0.898600 \\
        CITIC Capital Holdings & 0.898527 \\
        S3 Ventures & 0.898228 \\
        Chrysalix Venture Capital & 0.897628 \\
        BlueCross BlueShield Venture Partners & 0.897500 \\
        Foxconn Technology Group & 0.897093 \\
        VNV Global & 0.896948 \\
        Camden Partners & 0.896848 \\
        GrandBanks Capital & 0.896780 \\
        Rakuten & 0.896531 \\
        Lumira Ventures & 0.894433 \\
        Openspace & 0.894398 \\
        BRM Capital & 0.894394 \\
        VenturEast & 0.894311 \\
        Harmony Partners & 0.893963 \\
        Capital Today & 0.893536 \\
        Eastward Capital Partners & 0.893342 \\
        Takeda Ventures & 0.893278 \\
        Clarus Ventures & 0.893153 \\
        \hline
    \end{tabular}
  \end{minipage}%
  \begin{minipage}{.5\linewidth}
    \centering
        \caption{Founders scoring}
    \label{tab:founder_scores}
    \scriptsize
    \begin{tabular}{|c|c|}
        \hline
        \textbf{Name} & \textbf{Score} \\
        \hline
        Adam Neumann & 1.000000 \\
        David Fano & 0.997524 \\
        Abhi Maruvada & 0.985468 \\
        Emanuele Pagani & 0.981734 \\
        Ryan Norton & 0.979890 \\
        ... & ... \\
        Dr. Jonas Sundqvist & 0.089640 \\
        Onno Huyser & 0.084219 \\
        Stéphane Akaya & 0.084219 \\
        Mark Van Wyk & 0.084219 \\
        Pit Zens & 0.000000 \\
        \hline
    \end{tabular}
  \end{minipage}
\end{table}

\begin{table}[htbp]
    \centering
    \caption{Unicorns recommendation model}
    \label{tab:unicorn_scoring}
    \scriptsize
    \begin{tabular}{|c|c|c|}
        \hline
        \textbf{Name} & \textbf{Enter Series} & \textbf{Last Series Value} \\
        \hline
        Coursera & series\_c & 2 500 000 000 \\
        H2O.ai & series\_b & 1 700 000 000 \\
        Freee & series\_c & \\
        Cybereason & series\_c & 3 100 000 000 \\
        Honest Buildings & series\_b & \\
        Reserve & series\_unknown & \\
        Qumulo & series\_c & 1 200 000 000 \\
        Silent Circle & series\_c & \\
        Vlocity & series\_b & 1 000 000 000 \\
        Aura Financial & series\_c & \\
        MoneyLion & series\_a & \\
        Databricks & series\_c & 2 750 000 000 \\
        OakNorth & series\_b & 2 800 000 000 \\
        ProtectWise & series\_b & \\
        Earnest & series\_c & \\
        Cohesity & secondary\_market & 2 500 000 000 \\
        Spring & series\_c & \\
        Darktrace & series\_d & 1 650 000 000 \\
        Bread & series\_unknown & \\
        Guideline & series\_c & 1 150 000 000 \\
        Nexxiot & series\_b & \\
        Cato Networks & series\_c & 2 500 000 000 \\
        CloudMinds & series\_b & \\
        Stash Financial & series\_e & 1 400 000 000 \\
        Branch International & series\_c & \\
        Fungible & series\_c & 700 000 000 \\
        BlueVoyant & series\_b & \\
        Reali & series\_b & \\
        AMD Pensando & series\_c & \\
        IntSights & series\_d & \\
        BigID & series\_c & 1 250 000 000 \\
        Flutterwave & series\_b & 1 000 000 000 \\
        K Health & series\_c & 1 400 000 000 \\
        Doma & series\_c & \\
        BlockFi & series\_a & \\
        Tomorrow.io & series\_c & \\
        Kin Insurance & series\_b & \\
        TrueLayer & series\_c & 1 000 000 000 \\
        bitkey Japan & series\_unknown & \\
        MegazoneCloud & series\_b & \\
        \hline
    \end{tabular}
\end{table}

\end{document}